\documentclass{article}

\usepackage[english]{babel}
\usepackage[letterpaper,top=2cm,bottom=2cm,left=3cm,right=3cm]{geometry}
\usepackage{amsmath,graphicx,float,hyperref,algorithm,algorithmic}
\usepackage{tikz}
\usetikzlibrary{arrows.meta, positioning}
\usepackage{amsmath}
\usepackage{amssymb}

\title{STABLE: Gated Continual Learning for Large Language Models}

\author{
  William Hoy \qquad Nurcin Celik \\[0.5em]
  Department of Industrial and Systems Engineering, University of Miami \\[0.3em]
  \texttt{\{wjh58, celik\}@miami.edu}
}

\date{}

\begin{document}
\maketitle

\begin{abstract}
Large language models (LLMs) increasingly require mechanisms for continual adaptation
without full retraining. However, sequential updates can lead to catastrophic
forgetting, where new edits degrade previously acquired knowledge. This work presents STABLE, 
a gated continual self editing framework that constrains forgetting during
sequential updates using parameter efficient fine tuning (PEFT) via Low Rank Adaptation
(LoRA) \cite{hu2022lora}. Each candidate edit is evaluated against a stability budget
using one of three alternative metrics: (i) Exact Match (EM) drop, capturing factual
accuracy loss; (ii) Bits increase, reflecting reduced model confidence; and
(iii) KL divergence, quantifying distributional drift between the base and
adapted models. If a threshold is exceeded, the LoRA update is rescaled through a
clipping procedure or rejected.

Experiments on the Qwen-2.5-7B model show that gating effectively mitigates
forgetting while preserving adaptability. EM based gating achieved the highest 
cumulative performance in short continual learning sequences. Our results show that different gating strategies can achieve 
comparable distribution shift (measured by KL divergence) while producing different 
accuracy outcomes, highlighting the importance of gating design in continual adaptation. 
This approach offers a principled method for continual model editing, enabling LLMs to 
integrate new knowledge while maintaining reliability. The code is available at https://github.com/Bhoy1/STABLE. 
\end{abstract}

\section{Introduction}
Large language models (LLMs) have advanced the state of the art in tasks such as factual 
question answering, summarization, and dialogue generation. Their performance depends on 
large scale pretraining, but real world deployments increasingly require models that can 
be updated incrementally after release to serve as long term agents that continuously 
adapt to new needs. For instance, an LLM may need to incorporate new facts, domain 
specific terminology, or policy changes without undergoing full retraining. This motivates 
the study of continual editing, where models are repeatedly updated through lightweight 
interventions.

Despite progress in model editing, continual adaptation remains challenging. The main 
difficulty is catastrophic forgetting, the loss of previously acquired knowledge 
when new information is introduced. Forgetting reduces model reliability, particularly 
in long lived systems, where accumulated drift can propagate across tasks. 

Parameter efficient fine tuning (PEFT) methods such as Low Rank Adaptation (LoRA) 
\cite{hu2022lora} provide a practical mechanism for introducing new knowledge through 
compact trainable matrices while keeping most model parameters frozen. The SEAL 
framework \cite{zweiger2025seal} leverages LoRA adapters to enable models to 
autonomously propose self edits, improving factual accuracy over time. However, when 
these LoRA adapters are merged sequentially, earlier knowledge may degrade because 
successive updates can introduce unconstrained distributional shifts. 

Building on SEAL, our work addresses the complementary challenge of edit 
acceptance, determining whether a proposed update should be integrated into the base 
model without violating the gate's budget. This transforms SEAL’s open ended 
self editing process into a gated continual adaptation framework that enforces 
stability across sequential updates.

We present a gated continual self editing framework that constrains the updates during 
sequential LoRA merges. Each adapter merge is treated as a candidate update that must 
pass through a gate, which evaluates the edit against a user defined budget 
using one of three alternative metrics: (1) Exact Match (EM) drop, measuring loss in 
factual accuracy; (2) Bits increase, capturing reduced model confidence; and (3) KL 
divergence, quantifying distributional shift between the base and adapted models. The metric is chosen by the user. If the gate threshold is exceeded, the update is rescaled through a LoRA specific clipping 
procedure or rejected.

We evaluate the framework using the Qwen-2.5-7B model in a series of controlled editing 
experiments and find that gating mitigates catastrophic forgetting while 
maintaining adaptability. We situate these results within two relevant lines of work:  
(i) research on distribution shift in continual learning \cite{shenfeld2025rlrazor}, motivating the use of KL divergence as a proxy for catastrophic forgetting, and  
(ii) recent approaches to continual adaptation of language models \cite{zweiger2025seal}, where models incrementally integrate new knowledge during deployment.

\vspace{1.5em}
\noindent\textbf{Our primary contributions are threefold:}
\begin{itemize}
    \item A gating framework for continual LoRA adaptation with three alternative metrics for evaluating edits (based on the user's choosing): exact match (EM) for factual accuracy, 
    bits for model confidence, and KL divergence for distributional shift.
    
    \item Empirical comparison of gating strategies across all three metrics, showing that 
    metric choice significantly impacts the forgetting vs. adaption tradeoff. EM based gating 
    achieved the highest cumulative performance in our experiments.
    
    \item Demonstration that comparable KL divergence levels can yield different 
    task outcomes depending on the gating mechanism.
    
\end{itemize}

\section{Related Work}

Continual adaptation in large language models intersects several active research areas,
including catastrophic forgetting, parameter efficient fine tuning, self adaptive
editing, and constrained optimization. We review these domains below to contextualize
our proposed gated continual self editing framework.

\subsection{Catastrophic Forgetting}
Catastrophic forgetting is a longstanding challenge in continual learning, where 
models overwrite previously acquired knowledge when trained on new tasks 
\cite{mccloskey1989catastrophic,french1999catastrophic}. Classical strategies include 
rehearsal based replay \cite{rebuffi2017icarl,rolnick2019experience}, 
parameter isolation \cite{rusu2016progressive,fernando2017pathnet}, 
and regularization based methods such as Elastic Weight Consolidation (EWC) 
\cite{kirkpatrick2017ewc} and Synaptic Intelligence (SI) \cite{zenke2017continual}. 
These approaches protect prior knowledge by either reusing past data, allocating 
dedicated subnetworks, or penalizing deviation from weights deemed important for 
previous tasks \cite{wang2024survey}.

In the context of large language models, catastrophic forgetting has been observed 
during fine tuning, where adaptation to new tasks degrades general capabilities 
\cite{ouyang2022training,luo2023empirical}. Recent work has shown that KL divergence between the fine tuned and base model, 
measured on the new task, reliably predicts the degree of forgetting across 
different training algorithms \cite{shenfeld2025rlrazor}. This empirical forgetting 
law demonstrates that among multiple solutions, 
those with smaller KL divergence exhibit less catastrophic forgetting. This motivates 
our use of KL divergence as one of three gating metrics for controlling distributional 
drift during continual LoRA adaptation.

\subsection{PEFT}

Parameter efficient fine tuning (PEFT) methods aim to adapt large language models 
without updating all parameters. Low Rank Adaptation (LoRA) \cite{hu2022lora} inserts 
trainable rank decomposed matrices into transformer layers, yielding high performance 
while preserving the frozen base model. LoRA has become a standard tool for domain 
adaptation and continual fine tuning due to its efficiency and modularity.

Recent analysis by Thinking Machines Labs \cite{thinkingmachines2025lora} 
shows that LoRA fine tuning can closely match full supervised fine tuning (SFT) when 
the adapter has sufficient capacity relative to the data scale. This finding reinforces LoRA’s 
practicality for continual and post deployment adaptation, offering SFT performance 
at a fraction of the cost.

\subsection{Self Adaptive Language Models}
The SEAL framework \cite{zweiger2025seal} treats editing as a reinforcement 
learning problem, where the model proposes self edits and is rewarded based on 
improved factual accuracy and consistency. This paradigm moves beyond static 
fine tuning, enabling continual self improvement. However, in its current form, 
SEAL performs continual learning without an explicit gating mechanism to control forgetting. As a result, successive edits can overwrite 
previously learned information, leading to catastrophic forgetting over time. 
We find that this lack of gating represents a key limitation of SEAL's otherwise 
powerful self adaptive paradigm.

Our work builds on SEAL but extends it in a new direction. 
While SEAL focuses on the generation of candidate edits, 
we address the complementary challenge of acceptance control: 
ensuring that edits can be integrated without catastrophic forgetting.
We incorporate SEAL's self edit generation loop into our framework, 
but add a gating layer that evaluates each edit against gating metrics 
before acceptance. 

This combination yields a unified system in which SEAL proposes candidate edits 
and the gating mechanism filters or rescales them for stability, effectively 
bridging the gap between autonomous edit generation and reliable continual 
adaptation.

\subsection{Constrained Optimization in Model Updates}
The idea of bounding updates within constrained regions has analogues in reinforcement 
learning and language model alignment. KL regularization, which penalizes divergence 
between updated and reference policies, is a central mechanism for enforcing such 
constraints. 

Trust region methods operationalize this principle by gating policy updates based 
on measured divergence. Trust Region Policy Optimization (TRPO) \cite{schulman2015trpo} 
and Proximal Policy Optimization (PPO) \cite{schulman2017ppo} explicitly restrict 
updates whose KL distance exceeds a threshold, ensuring stable optimization. We adopt 
the same principle in continual editing, where adapter merges are treated analogously 
to bounded policy updates.

\section{Methodology}

\subsection{Gated Continual Self Editing}

Our implementation builds on the open source SEAL framework \cite{zweiger2025seal}, 
which trains language models to propose self edits via reinforcement learning. 
While SEAL focuses on edit generation, we extend its codebase with a gated 
acceptance control layer. In our setting, edits are provided as LoRA modules and 
must pass through a gate before being merged. This gate enforces a 
tunable budget, measured via one of three metrics: 
exact match (EM) drop, bits increase, or KL divergence. If the budget is violated, 
the LoRA weights are rescaled through clipping or the edit is rejected outright. 
This design preserves SEAL's self adaptive editing capabilities while adding 
explicit control over knowledge retention.

\usetikzlibrary{arrows.meta, positioning, calc, shadows, decorations.pathreplacing}

\begin{figure}[H]
\centering
\resizebox{0.9\textwidth}{!}{%
\begin{tikzpicture}[
  block/.style = {
    rectangle, 
    draw=blue!60, 
    fill=blue!5,
    line width=0.8pt,
    rounded corners=2pt,
    minimum width=2.2cm, 
    minimum height=1cm, 
    align=center,
    drop shadow={opacity=0.15, shadow xshift=0.5pt, shadow yshift=-0.5pt}
  },
  gateblock/.style = {
    rectangle, 
    draw=orange!70, 
    fill=orange!10,
    line width=1pt,
    rounded corners=2pt,
    minimum width=2.2cm, 
    minimum height=1cm, 
    align=center,
    drop shadow={opacity=0.2, shadow xshift=0.5pt, shadow yshift=-0.5pt}
  },
  arrow/.style = {
    thick, 
    -{Stealth[length=3mm]},
    draw=gray!70
  },
  title/.style = {
    font=\bfseries\large,
    align=center
  }
]

% --- Top row: Unconstrained ---
\node (base1) [block] {Base model \\ \(\theta_{\text{base}}\)};
\node (edit1) [block, right=of base1, xshift=1cm] {Propose LoRA edit \\ \(W_{\text{LoRA}}\)};
\node (merge1) [block, right=of edit1, xshift=1cm] {Merge without gate \\ \(\to \theta_{\text{new}}\)};
\node (eval1) [block, right=of merge1, xshift=1cm] {Evaluate forgetting};

\draw[arrow] (base1) -- (edit1);
\draw[arrow] (edit1) -- (merge1);
\draw[arrow] (merge1) -- (eval1);

% --- Bottom row: Gated ---
\node (base2) [block, below=of base1, yshift=-2cm] {Base model \\ \(\theta_{\text{base}}\)};
\node (edit2) [block, right=of base2, xshift=1cm] {Propose LoRA edit \\ \(W_{\text{LoRA}}\)};
\node (gate)  [gateblock, right=of edit2, xshift=1cm] {%
  \textbf{Gate} \\ 
  compute \(f\), \\ 
  check \(f \le \epsilon\)%
};
\node (merge2) [block, right=of gate, xshift=1cm] {Merge (or scale/reject) \\ \(\to \theta_{\text{new}}\)};
\node (eval2) [block, right=of merge2, xshift=1cm] {Evaluate forgetting};

\draw[arrow] (base2) -- (edit2);
\draw[arrow] (edit2) -- (gate);
\draw[arrow] (gate) -- (merge2);
\draw[arrow] (merge2) -- (eval2);

% --- Calculate the true center of the entire figure ---
\coordinate (figcenter) at ($(base1)!0.5!(eval1)$);

% --- Titles aligned to figure center ---
\node[title, anchor=south] at ([xshift=1.8cm, yshift=0.8cm]figcenter |- base1.north) {Unconstrained Merge};
\node[title, anchor=south] at ([xshift=1.8cm, yshift=0.8cm]figcenter |- base2.north) {Gated Editing};

% --- Optional: Highlight key difference with brace ---
\draw[decorate, decoration={brace, amplitude=8pt, mirror}, 
      thick, orange!70] 
      ([yshift=-0.3cm]gate.south west) -- 
      ([yshift=-0.3cm]gate.south east)
      node[midway, below=8pt, align=center, font=\small\itshape, text=orange!80] 
      {Key difference};

\end{tikzpicture}%
}
\caption{Comparison of unconstrained LoRA merging (top) versus gated merging (bottom). The gate ensures edits are validated for forgetting before merges are accepted, providing a critical safeguard against performance degradation.}
\label{fig:edit_pipeline_wide}
\end{figure}
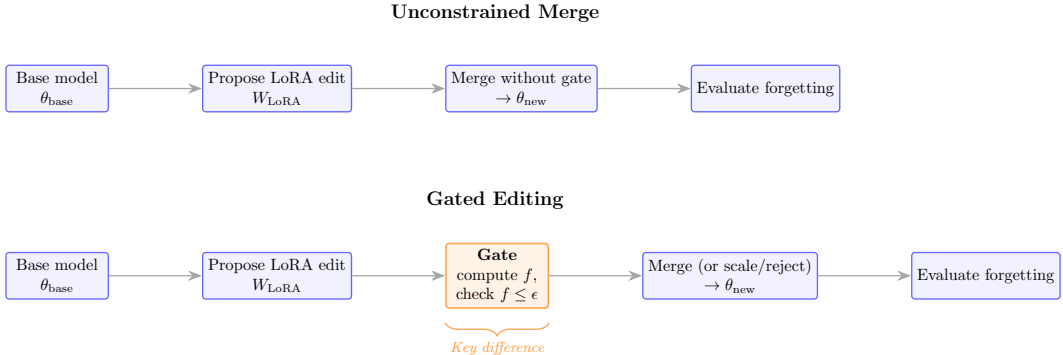

\subsubsection{Formulation}
The gating mechanism can be formalized as a constraint on how much an edit is allowed 
to alter prior knowledge. We express this constraint mathematically as follows.

Let $f$ denote a forgetting measure and $\epsilon$ a budget. 
A merge is accepted only if
\[
f \leq \epsilon,
\]
otherwise the edit is either rejected or rescaled by a factor 
$\alpha \in [\alpha_{\min}, 1]$ to satisfy the constraint. 
We compare three alternative metrics for $f$.

\paragraph{Exact Match (EM) Drop.}
The EM metric quantifies task level forgetting as the reduction in factual accuracy on
anchor questions after adaptation. For each question $q_i$ with gold answer $y_i$, a
large language model grader provides a binary reward
\[
\text{LLM}(q_i, \hat{y}) \in \{0,1\},
\]
which equals $1$ if the generated answer $\hat{y}$ matches $y_i$, and $0$ otherwise.
Baseline and adapter accuracies are computed as
\[
\text{EM}_{\text{base}} = \frac{1}{N}\sum_{i=1}^N \text{LLM}(q_i, \hat{y}_i^{\text{base}}),
\quad
\text{EM}_{\text{adapter}} = \frac{1}{N}\sum_{i=1}^N \text{LLM}(q_i, \hat{y}_i^{\text{adapter}}).
\]
The forgetting signal is then defined as
\[
f_{\mathrm{EM}} = \max\!\left(0,\, \text{EM}_{\text{base}} - \text{EM}_{\text{adapter}}\right).
\]
This metric captures the portion of factual accuracy lost after merging.
For example, $f_{\mathrm{EM}} = 0.05$ indicates a $5\%$ decline in correctness on anchor
questions, while $f_{\mathrm{EM}} = 0.10$ represents a moderate $10\%$ degradation in
factual retention.

\paragraph{Bits Increase.}
The bits metric measures confidence level forgetting by comparing how efficiently each
model encodes its own answers, expressed in bits per token. Both the base and adapter
models independently generate answers to the same anchor questions and then evaluate the
log probabilities of their own outputs. A less confident model assigns lower probability
to its generated tokens, resulting in a higher average bits per token score.

For a sequence $x_{1:T}$ produced by model $\theta$, we define
\[
\text{bits}(\theta) = - \frac{1}{T \log 2} \sum_{t=1}^T \log p_\theta(x_t),
\]
and quantify forgetting as
\[
f_{\mathrm{bits}} = 
\max\!\left(0,\;
\text{bits}_{\text{adapter}} -
\text{bits}_{\text{base}}
\right),
\]
where $\text{bits}$ denotes the mean across all anchor questions.
This metric reflects how much additional information (in bits) the adapter model requires
to express the same knowledge. Even if both models produce identical answers, a higher
bits score indicates reduced confidence. For instance, a 0.08 bits/token threshold requires the adapter to remain approximately 
95\% as confident as the base model on average 
\[
\frac{p_\mathrm{adapter}}{p_\mathrm{base}} = 2^{-\Delta_\mathrm{bits}}.
\]

\paragraph{KL Divergence.}
We compute the KL divergence between the base (reference) model and the adapted
(LoRA merged) model using the log probabilities of the generated tokens themselves. 
For each token $x_t$ in a sampled sequence, let $\pi_{\text{base}}$ and 
$\pi_{\text{adapter}}$ denote the conditional distributions of the base and adapter 
models, respectively. The per token KL term is
\[
D_{\mathrm{KL}}\!\left(\pi_{\text{adapter}}(\cdot \mid x_{<t}) \,\|\, 
\pi_{\text{base}}(\cdot \mid x_{<t})\right)
= \mathbb{E}_{x_t \sim \pi_{\text{adapter}}}
\!\left[\log_2 
\frac{\pi_{\text{adapter}}(x_t \mid x_{<t})}
     {\pi_{\text{base}}(x_t \mid x_{<t})}\right].
\]
The overall drift metric averages this divergence over all tokens:
\[
f_{\mathrm{KL}} = 
\frac{1}{T \log 2}
\sum_{t=1}^{T}
\left(
\log \pi_{\text{adapter}}(x_t \mid x_{<t})
-
\log \pi_{\text{base}}(x_t \mid x_{<t})
\right),
\]
where the division by $\log 2$ converts from nats to bits per token.
KL measures distributional drift between the two models, the extent to which the 
adapter's token probabilities diverge from the base model. A KL of 0 indicates identical distributions, while higher values 
reflect increasing divergence. Lower KL values indicate stronger alignment with 
the base model and reduced distributional drift.

\subsection{Gate Logic}
Our framework employs a LoRA gating mechanism, which ensures
each adapter merge satisfies the forgetting constraint. If the computed forgetting
score $f$ exceeds the budget $\epsilon$, a binary search is performed over the scaling
factor $\alpha \in [\alpha_{\min}, 1]$ to find the largest value that satisfies
\[
f(\alpha \cdot W_{\text{LoRA}}) \leq \epsilon,
\]
where $W_{\text{LoRA}}$ denotes the LoRA weight update. If no such $\alpha$ exists 
above the minimum threshold $\alpha_{\min}$, the merge is rejected.  

This procedure effectively projects edits onto the feasible region of bounded
forgetting, ensuring that all accepted merges adhere to the defined stability
constraint.

\begin{algorithm}[H]
\caption{Gated LoRA Merge (LoRA Clip)}
\begin{algorithmic}[1]
\STATE \textbf{Input:} base model $\theta$, LoRA update $W_{\text{LoRA}}$, budget $\epsilon$, min scale $\alpha_{\min}$
\STATE Compute forgetting score $f(W_{\text{LoRA}})$
\IF{$f(W_{\text{LoRA}}) \leq \epsilon$}
    \STATE \textbf{Accept} full merge with $\alpha = 1$
\ELSE
    \STATE Binary search for $\alpha \in [\alpha_{\min}, 1]$ satisfying $f(\alpha \cdot W_{\text{LoRA}}) \leq \epsilon$
    \IF{such $\alpha \geq \alpha_{\min}$ exists}
        \STATE \textbf{Accept} scaled merge with factor $\alpha$
    \ELSE
        \STATE \textbf{Reject} merge
    \ENDIF
\ENDIF
\STATE \textbf{Output:} Accept/Reject decision and scaling factor (if accepted)
\end{algorithmic}
\end{algorithm}

\section{Results and Analysis}

We conducted continual editing experiments on the Qwen2.5-7B model comparing three 
gating strategies, each based on a different forgetting metric: EM based gating 
(7\% threshold), bits based gating (0.08 bits/token threshold), and KL based gating 
(0.7 bits/token threshold). Each gating strategy was evaluated independently across 
12 runs, where each run randomly sampled 8 datapoints from the main SQuAD style 
dataset. As the sequence of self edits progresses, previously edited datapoints 
become anchors used to measure forgetting on past knowledge. This setup ensures that 
every run experiences a distinct sequence of edits and anchor configurations, 
enabling robust estimates of average performance trends and variability under 
stochastic sampling.

Two evaluations were conducted 
for each gating strategy under identical experimental conditions. The results 
presented in this section correspond to the best performing evaluation for each gate, 
selected based on step and cumulative performance. Complete results 
for both evaluations, including all intermediate statistics, are provided in 
Appendix~C. 

\begin{figure}[H]
    \centering
    \includegraphics[width=0.8\textwidth]{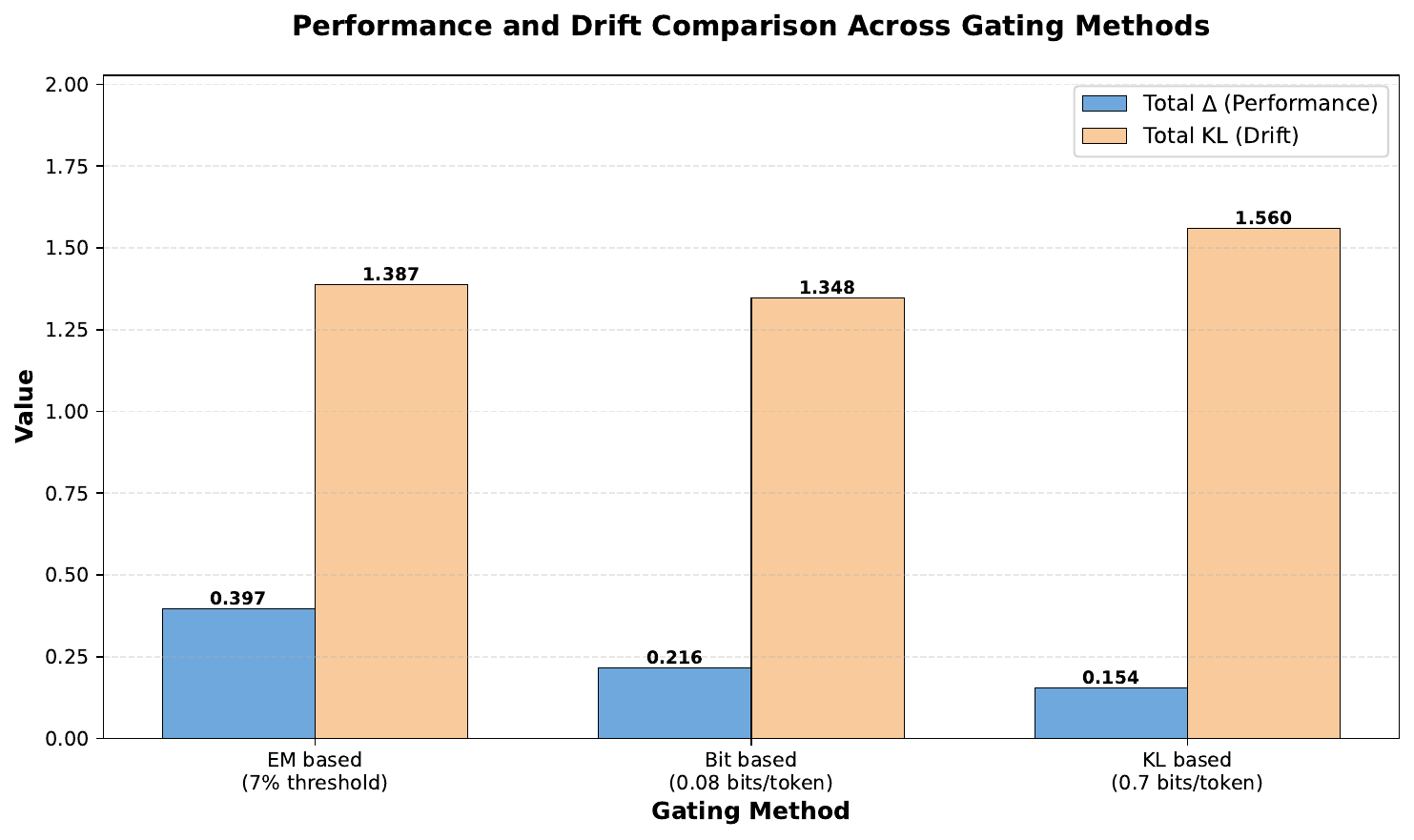}
    \caption{Performance and drift comparison across gating methods. EM based gating achieves the highest cumulative performance gain (+0.397).}
    \label{fig:gating_comparison}
\end{figure}

\begin{figure}[H]
    \centering
    \includegraphics[width=0.85\textwidth]{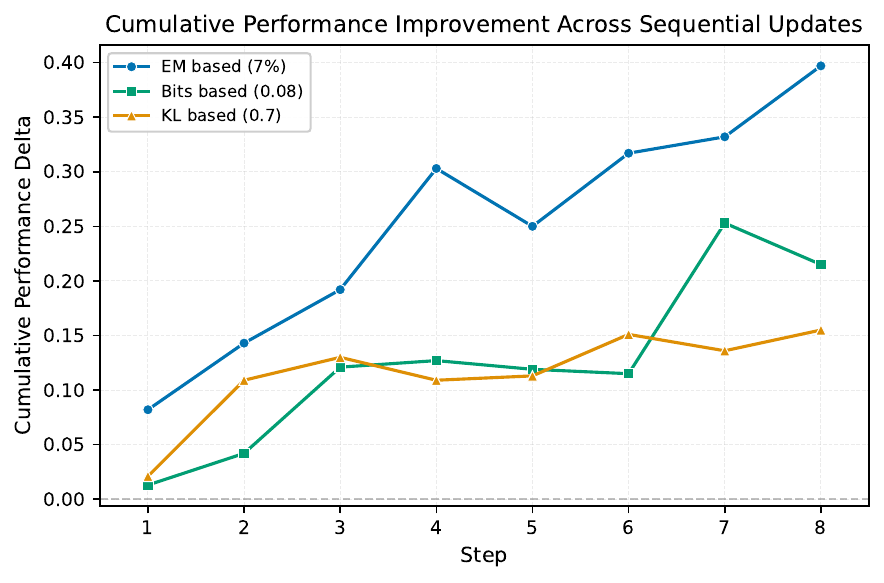}
    \caption{Cumulative performance improvement across sequential updates for the three 
    gating methods. EM based gating shows best results per step and cumulative.}
    \label{fig:cumulative_delta}
\end{figure}

\autoref{fig:cumulative_delta} illustrates the cumulative performance trajectory across 
sequential updates, showing how EM based gating performed the best across each step, while KL based and bit based methods maintain 
similar but ultimately lower cumulative improvements.

\begin{figure}[H]
    \centering
    \includegraphics[width=0.85\textwidth]{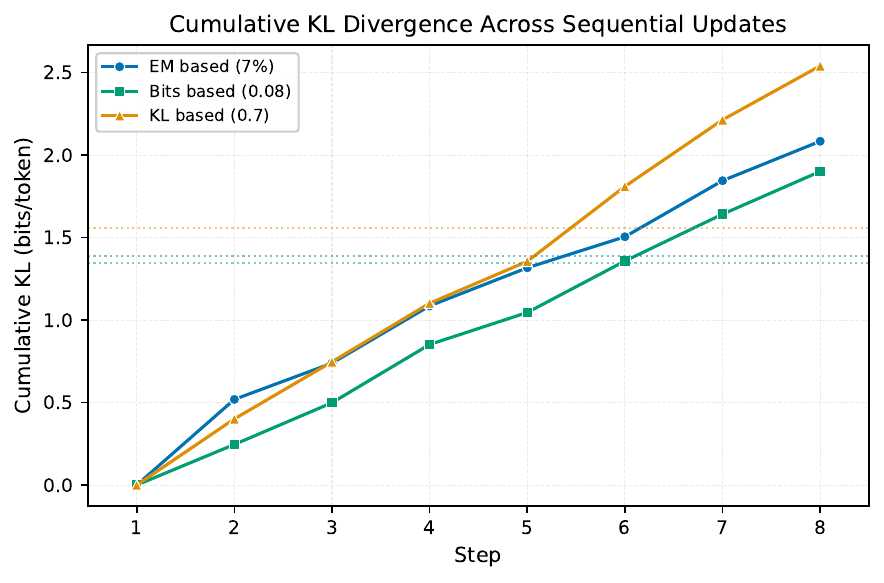}
    \caption{Cumulative KL divergence across sequential updates for the three gating methods. Dashed horizontal lines indicate the overall KL from baseline to final adapted model, which differs from the sum of per step KL values.}
    \label{fig:cumulative_kl}
\end{figure}

Per step KL divergence values followed the same ordering as overall KL, with all three metrics accumulating 
steadily across the eight sequential updates (\autoref{fig:cumulative_kl}). 
This demonstrates that gating strategy significantly impacts task performance even 
when distributional shift is comparable across approaches.

\section{Discussion}
\subsection{Theoretical Interpretation}
The gating framework can be interpreted through the lens of constrained optimization. 
Each LoRA merge represents a candidate update that may improve alignment with new knowledge 
but also risks degrading previously consolidated facts. The gate functions as a projection 
operator: if an update exceeds the feasible region defined by the forgetting budget 
$\epsilon$, it is scaled back to satisfy the constraint or rejected entirely. This mirrors 
safe reinforcement learning methods, where updates are constrained by KL divergence 
penalties to preserve policy stability. By introducing explicit feasibility checks, 
the framework transforms continual editing from an unconstrained process prone to drift 
into a controlled adaptation procedure that enforces stability while enabling knowledge 
integration.

\subsection{Limitations}
Despite its strengths, the proposed approach has several limitations. 
First, gating introduces additional computational overhead, as each candidate merge 
requires evaluating the gating metric before acceptance. While this cost is manageable 
in small scale experiments, it may become significant in large scale deployments. 
Second, the strictness of the gates can sometimes reject edits that, while slightly 
exceeding the budget, may have been beneficial overall. This rigidity illustrates an 
inherent tradeoff between remembering and adaptability. Finally, the current evaluation 
focuses on SQuAD style factual editing tasks. Although these tasks provide a controlled 
testbed, broader benchmarks spanning diverse domains and task types are needed to fully 
assess the method's generality.

\subsection{Future Work}
This study demonstrates that gating and SEAL style self edit generation can be combined 
into a unified framework, enabling models that both propose and evaluate their own edits. 
Building on this foundation, several extensions remain open. One direction is the 
development of adaptive forgetting budgets that adjust dynamically based on task importance 
or context, rather than remaining fixed across all edits. Another is scaling to real time 
and streaming environments, where efficiency and robustness under continuous inputs become 
critical. Together, these advances would move gated self editing closer to production ready continual 
adaptation.

\section{Conclusion}
This paper introduced gated continual self editing as a principled approach 
to mitigate catastrophic forgetting in LoRA based large language models. The motivation 
stems from the growing need for models that can be incrementally updated after 
deployment without erasing previously consolidated knowledge. Sequential adapter 
merges, while efficient, can accumulate instability if left unconstrained, motivating 
the need for explicit gating.

We proposed a framework in which each candidate edit is evaluated against a gating 
budget using one of three alternative metrics: exact match (EM) drop, bits increase, 
and KL divergence. If the budget is violated, the merge is either rescaled through 
LoRA clipping or rejected outright. This transforms the editing process into a 
constrained optimization problem, ensuring bounded forgetting.

Empirical evaluation on SQuAD style factual editing tasks demonstrated that different 
gating metrics produce qualitatively different adaptation dynamics. EM based gating 
achieved the highest cumulative improvement (40\%) in our eight step experiments. KL based and bits based 
both had lower performance. Notably, all three metrics 
achieved comparable distributional drift, demonstrating that gating strategy impacts 
task performance independently of drift control.

Overall, gating provides a practical and theoretically grounded pathway toward 
controlled continual adaptation of LLMs. By bounding forgetting through tunable constraints, 
it enables models to evolve incrementally while retaining prior knowledge. This work 
represents a step toward more reliable, long lived language models and motivates future 
extensions that integrate gating with self directed editing, adaptive budget allocation, 
and real time deployment in production environments.

\section*{Acknowledgments}
We gratefully acknowledge the authors of SEAL \cite{zweiger2025seal} for releasing 
their codebase, which served as the foundation for our implementation. 

\bibliographystyle{ieeetr}
\bibliography{refs}

\appendix

\section{Compute Resources}
All experiments were conducted on two NVIDIA A100 GPUs (80GB memory each).  
Training runs for the full Qwen-2.5-7B experiments required approximately 
20 hours wall clock time.  

\section{Training Parameters}
\begin{itemize}
    \item \textbf{LoRA configuration:} rank 32, scaling factor $\alpha=64$,
    trained for 10 epochs with a learning rate of $1\text{e-}3$. Only LoRA parameters were updated.  
    \item \textbf{Batching:} batch size 1
    \item \textbf{Binary search passes:} 5 evaluations per edit for LoRA clipping.  
    \item \textbf{Evaluation passes:} 12 passes across 8 datapoints    
    \item \textbf{Safety mode:} LoRA clip with minimum scale factor 0.1.
    \item 
    \textbf{Seed:} 53
\end{itemize}

\section{Additional Results}

\subsection{EM Based Gating}

\begin{table}[H]
\centering
\begin{tabular}{c|c|c|c|c|c}
\hline
Threshold & Step & Baseline ($\pm$ SE) & Final ($\pm$ SE) & $\Delta$ & KL (mean $\pm$ SE) \\
\hline
10\% & 1 & 0.222 $\pm$ 0.070 & 0.179 $\pm$ 0.067 & -0.043 & -- \\
     & 2 & 0.325 $\pm$ 0.084 & 0.275 $\pm$ 0.080 & -0.050 & 0.352 $\pm$ 0.072 \\
     & 3 & 0.364 $\pm$ 0.070 & 0.346 $\pm$ 0.073 & -0.018 & 0.390 $\pm$ 0.202 \\
     & 4 & 0.338 $\pm$ 0.066 & 0.369 $\pm$ 0.052 & +0.032 & 0.476 $\pm$ 0.203 \\
     & 5 & 0.306 $\pm$ 0.049 & 0.383 $\pm$ 0.075 & +0.078 & 0.278 $\pm$ 0.065 \\
     & 6 & 0.367 $\pm$ 0.063 & 0.433 $\pm$ 0.078 & +0.067 & 0.261 $\pm$ 0.073 \\
     & 7 & 0.272 $\pm$ 0.062 & 0.433 $\pm$ 0.080 & +0.161 & 0.335 $\pm$ 0.062 \\
     & 8 & 0.503 $\pm$ 0.063 & 0.605 $\pm$ 0.075 & +0.103 & 0.372 $\pm$ 0.078 \\
\hline
\textbf{10\%} & \textbf{Total} & \textbf{2.697} & \textbf{3.023} & \textbf{+0.330} & \textbf{1.457 $\pm$ 0.085} \\
\hline
7\% & 1 & 0.201 $\pm$ 0.066 & 0.283 $\pm$ 0.095 & +0.082 & -- \\
    & 2 & 0.325 $\pm$ 0.074 & 0.386 $\pm$ 0.095 & +0.061 & 0.519 $\pm$ 0.215 \\
    & 3 & 0.381 $\pm$ 0.077 & 0.429 $\pm$ 0.091 & +0.049 & 0.220 $\pm$ 0.064 \\
    & 4 & 0.358 $\pm$ 0.067 & 0.469 $\pm$ 0.078 & +0.111 & 0.347 $\pm$ 0.091 \\
    & 5 & 0.322 $\pm$ 0.049 & 0.269 $\pm$ 0.070 & -0.053 & 0.231 $\pm$ 0.057 \\
    & 6 & 0.346 $\pm$ 0.063 & 0.413 $\pm$ 0.060 & +0.067 & 0.188 $\pm$ 0.067 \\
    & 7 & 0.293 $\pm$ 0.057 & 0.308 $\pm$ 0.055 & +0.015 & 0.340 $\pm$ 0.104 \\
    & 8 & 0.453 $\pm$ 0.075 & 0.518 $\pm$ 0.070 & +0.065 & 0.239 $\pm$ 0.050 \\
\hline
\textbf{7\%} & \textbf{Total} & \textbf{2.679} & \textbf{3.076} & \textbf{+0.397} & \textbf{1.387 $\pm$ 0.093} \\
\hline
\end{tabular}
\caption{Performance under EM based gating at 10\% and 7\% thresholds. 
Baseline: base model with no continual learning. 
Final: adapted model after 8 sequential updates. 
KL divergence shows per step drift between consecutive models (starting at step 2). 
Total KL represents the global drift from base to final model.}
\label{tab:em_results}
\end{table}

The table below summarizes acceptance and scaling behavior across all passes.

\begin{table}[H]
\centering
\begin{tabular}{c|c|r|r|r|r}
\hline
Threshold & Step & Accept & Scaled & Rejected & Mean Scale ($\pm$SE) \\
\hline
10\% & \textbf{2} & 6 & 6 & 0 & $\mathbf{0.812 \pm 0.068}$ \\
     & \textbf{3} & 6 & 6 & 0 & $\mathbf{0.784 \pm 0.077}$ \\
     & \textbf{4} & 7 & 5 & 0 & $\mathbf{0.885 \pm 0.060}$ \\
     & \textbf{5} & 9 & 3 & 0 & $\mathbf{0.909 \pm 0.055}$ \\
     & \textbf{6} & 6 & 6 & 0 & $\mathbf{0.838 \pm 0.065}$ \\
     & \textbf{7} & 9 & 3 & 0 & $\mathbf{0.932 \pm 0.044}$ \\
     & \textbf{8} & 8 & 4 & 0 & $\mathbf{0.930 \pm 0.047}$ \\
\hline
7\% & \textbf{2} & 6 & 6 & 0 & $\mathbf{0.794 \pm 0.079}$ \\
    & \textbf{3} & 4 & 8 & 0 & $\mathbf{0.829 \pm 0.059}$ \\
    & \textbf{4} & 6 & 6 & 0 & $\mathbf{0.852 \pm 0.069}$  \\
    & \textbf{5} & 8 & 4 & 0 & $\mathbf{0.852 \pm 0.074}$ \\
    & \textbf{6} & 6 & 6 & 0 & $\mathbf{0.768 \pm 0.078}$ \\
    & \textbf{7} & 7 & 5 & 0 & $\mathbf{0.883 \pm 0.060}$ \\
    & \textbf{8} & 6 & 6 & 0 & $\mathbf{0.883 \pm 0.043}$ \\
\hline
\end{tabular}
\caption{Per step acceptance, scaling, and rejection statistics for EM based gating
at 10\% and 7\% thresholds. Step 1 is omitted as no scaling is applied.}
\label{tab:appendix_em_gating}
\end{table}

\subsection{Bits Based Gating}

\begin{table}[H]
\centering
\begin{tabular}{c|c|c|c|c|c}
\hline
Threshold & Step & Baseline ($\pm$ SE) & Final ($\pm$ SE) & $\Delta$ & KL (mean $\pm$ SE) \\
\hline
0.08 & 1 & 0.201 $\pm$ 0.066 & 0.214 $\pm$ 0.067 & +0.013 & -- \\
     & 2 & 0.308 $\pm$ 0.074 & 0.338 $\pm$ 0.089 & +0.029 & 0.246 $\pm$ 0.057 \\
     & 3 & 0.347 $\pm$ 0.067 & 0.426 $\pm$ 0.080 & +0.079 & 0.253 $\pm$ 0.067 \\
     & 4 & 0.338 $\pm$ 0.066 & 0.344 $\pm$ 0.072 & +0.006 & 0.353 $\pm$ 0.119 \\
     & 5 & 0.301 $\pm$ 0.046 & 0.293 $\pm$ 0.079 & -0.008 & 0.194 $\pm$ 0.054 \\
     & 6 & 0.346 $\pm$ 0.063 & 0.342 $\pm$ 0.055 & -0.004 & 0.312 $\pm$ 0.080 \\
     & 7 & 0.234 $\pm$ 0.048 & 0.373 $\pm$ 0.061 & +0.138 & 0.284 $\pm$ 0.068 \\
     & 8 & 0.498 $\pm$ 0.072 & 0.460 $\pm$ 0.074 & -0.038 & 0.259 $\pm$ 0.064 \\
\hline
\textbf{0.08} & \textbf{Total} & \textbf{2.574} & \textbf{2.789} & \textbf{+0.216} & \textbf{1.348 $\pm$ 0.067} \\
\hline
0.06 & 1 & 0.201 $\pm$ 0.066 & 0.210 $\pm$ 0.081 & +0.008 & -- \\
     & 2 & 0.308 $\pm$ 0.074 & 0.392 $\pm$ 0.079 & +0.083 & 0.221 $\pm$ 0.086 \\
     & 3 & 0.364 $\pm$ 0.074 & 0.421 $\pm$ 0.076 & +0.057 & 0.855 $\pm$ 0.353 \\
     & 4 & 0.392 $\pm$ 0.068 & 0.296 $\pm$ 0.068 & -0.096 & 0.176 $\pm$ 0.072 \\
     & 5 & 0.306 $\pm$ 0.049 & 0.383 $\pm$ 0.061 & +0.078 & 0.052 $\pm$ 0.016 \\
     & 6 & 0.350 $\pm$ 0.060 & 0.375 $\pm$ 0.080 & +0.025 & 0.144 $\pm$ 0.040 \\
     & 7 & 0.255 $\pm$ 0.066 & 0.222 $\pm$ 0.063 & -0.033 & 0.148 $\pm$ 0.044 \\
     & 8 & 0.486 $\pm$ 0.058 & 0.510 $\pm$ 0.085 & +0.024 & 0.156 $\pm$ 0.050 \\
\hline
\textbf{0.06} & \textbf{Total} & \textbf{2.662} & \textbf{2.809} & \textbf{+0.147} & \textbf{1.835 $\pm$ 0.640} \\
\hline
\end{tabular}
\caption{Performance under bits based gating at 0.08 and 0.06 bits/token thresholds. 
Baseline: base model with no continual learning. 
Final: adapted model after 8 sequential updates. 
KL divergence shows per step drift between consecutive models (starting at step 2). 
Total KL represents the drift from base to final model.}
\label{tab:bits_results}
\end{table}

The table below summarizes acceptance and scaling behavior across all passes.

\begin{table}[H]
\centering
\begin{tabular}{c|c|r|r|r|r}
\hline
Threshold & Step & Accept & Scaled & Rejected & Mean Scale ($\pm$SE) \\
\hline
0.08 & \textbf{2} & 1 & 11 & 0 & $\mathbf{0.714 \pm 0.065}$ \\
     & \textbf{3} & 4 & 8 & 0 & $\mathbf{0.841 \pm 0.053}$ \\
     & \textbf{4} & 7 & 5 & 0 & $\mathbf{0.822 \pm 0.069}$ \\
     & \textbf{5} & 4 & 8 & 0 & $\mathbf{0.827 \pm 0.063}$ \\
     & \textbf{6} & 6 & 6 & 0 & $\mathbf{0.848 \pm 0.060}$ \\
     & \textbf{7} & 5 & 7 & 0 & $\mathbf{0.810 \pm 0.063}$ \\
     & \textbf{8} & 7 & 5 & 0 & $\mathbf{0.937 \pm 0.044}$ \\
\hline
0.06 & \textbf{2} & 5 & 7 & 0 & $\mathbf{0.733 \pm 0.091}$ \\
     & \textbf{3} & 5 & 7 & 0 & $\mathbf{0.766 \pm 0.092}$ \\
     & \textbf{4} & 5 & 7 & 0 & $\mathbf{0.693 \pm 0.089}$ \\
     & \textbf{5} & 3 & 9 & 0 & $\mathbf{0.644 \pm 0.078}$ \\
     & \textbf{6} & 1 & 9 & 2 & $\mathbf{0.798 \pm 0.050}$ \\
     & \textbf{7} & 2 & 10 & 0 & $\mathbf{0.691 \pm 0.067}$ \\
     & \textbf{8} & 2 & 10 & 0 & $\mathbf{0.702 \pm 0.070}$ \\
\hline
\end{tabular}
\caption{Per step acceptance, scaling, and rejection statistics for bits based gating
at 0.08 and 0.06 bits per token thresholds. Step 1 is omitted as no scaling 
is applied.}
\label{tab:appendix_bits_gating}
\end{table}

\subsection{KL Based Gating}

\begin{table}[H]
\centering
\begin{tabular}{c|c|c|c|c|c}
\hline
Threshold & Step & Baseline ($\pm$ SE) & Final ($\pm$ SE) & $\Delta$ & KL (mean $\pm$ SE) \\
\hline
0.7 & 1 & 0.201 $\pm$ 0.066 & 0.222 $\pm$ 0.064 & +0.021 & -- \\
    & 2 & 0.325 $\pm$ 0.074 & 0.413 $\pm$ 0.069 & +0.088 & 0.401 $\pm$ 0.053 \\
    & 3 & 0.364 $\pm$ 0.074 & 0.385 $\pm$ 0.083 & +0.021 & 0.346 $\pm$ 0.074 \\
    & 4 & 0.338 $\pm$ 0.066 & 0.317 $\pm$ 0.062 & -0.021 & 0.355 $\pm$ 0.061 \\
    & 5 & 0.318 $\pm$ 0.046 & 0.322 $\pm$ 0.055 & +0.004 & 0.254 $\pm$ 0.053 \\
    & 6 & 0.350 $\pm$ 0.060 & 0.388 $\pm$ 0.082 & +0.038 & 0.454 $\pm$ 0.061 \\
    & 7 & 0.255 $\pm$ 0.049 & 0.240 $\pm$ 0.070 & -0.015 & 0.404 $\pm$ 0.053 \\
    & 8 & 0.453 $\pm$ 0.071 & 0.471 $\pm$ 0.081 & +0.019 & 0.327 $\pm$ 0.058 \\
\hline
\textbf{0.7} & \textbf{Total} & \textbf{2.604} & \textbf{2.758} & \textbf{+0.154} & \textbf{1.560 $\pm$ 0.083} \\
\hline
0.5 & 1 & 0.201 $\pm$ 0.066 & 0.188 $\pm$ 0.060 & -0.014 & -- \\
    & 2 & 0.308 $\pm$ 0.074 & 0.308 $\pm$ 0.095 & +0.000 & 0.343 $\pm$ 0.043 \\
    & 3 & 0.336 $\pm$ 0.076 & 0.239 $\pm$ 0.068 & -0.097 & 0.268 $\pm$ 0.059 \\
    & 4 & 0.342 $\pm$ 0.073 & 0.258 $\pm$ 0.056 & -0.083 & 0.308 $\pm$ 0.050 \\
    & 5 & 0.285 $\pm$ 0.046 & 0.336 $\pm$ 0.094 & +0.051 & 0.338 $\pm$ 0.040 \\
    & 6 & 0.350 $\pm$ 0.060 & 0.329 $\pm$ 0.088 & -0.021 & 0.345 $\pm$ 0.034 \\
    & 7 & 0.251 $\pm$ 0.049 & 0.294 $\pm$ 0.067 & +0.044 & 0.383 $\pm$ 0.033 \\
    & 8 & 0.486 $\pm$ 0.068 & 0.405 $\pm$ 0.091 & -0.081 & 0.321 $\pm$ 0.036 \\
\hline
\textbf{0.5} & \textbf{Total} & \textbf{2.559} & \textbf{2.358} & \textbf{-0.201} & \textbf{1.458 $\pm$ 0.053} \\
\hline
\end{tabular}
\caption{Performance under KL based gating at 0.7 and 0.5 bits/token thresholds. 
Baseline: base model with no continual learning. 
Final: adapted model after 8 sequential updates. 
KL divergence shows per step drift between consecutive models (starting at step 2). 
Total KL represents the drift from base to final model.}
\label{tab:kl_results}
\end{table}

The table below summarizes acceptance and scaling behavior across all passes.

\begin{table}[H]
\centering
\begin{tabular}{c|c|r|r|r|r}
\hline
Threshold & Step & Accept & Scaled & Rejected & Mean Scale ($\pm$SE) \\
\hline
0.7 & \textbf{2} & 4 & 5 & 3 & $\mathbf{0.951 \pm 0.022}$ \\
    & \textbf{3} & 8 & 2 & 2 & $\mathbf{0.988 \pm 0.008}$ \\
    & \textbf{4} & 9 & 1 & 2 & $\mathbf{0.991 \pm 0.009}$ \\
    & \textbf{5} & 10 & 1 & 1 & $\mathbf{0.995 \pm 0.005}$ \\
    & \textbf{6} & 8 & 3 & 1 & $\mathbf{0.972 \pm 0.019}$ \\
    & \textbf{7} & 10 & 2 & 0 & $\mathbf{0.986 \pm 0.009}$ \\
    & \textbf{8} & 11 & 1 & 0 & $\mathbf{0.993 \pm 0.007}$ \\
\hline
0.5 & \textbf{2} & 5 & 4 & 3 & $\mathbf{0.956 \pm 0.020}$ \\
    & \textbf{3} & 6 & 3 & 3 & $\mathbf{0.970 \pm 0.018}$ \\
    & \textbf{4} & 6 & 3 & 3 & $\mathbf{0.984 \pm 0.010}$ \\
    & \textbf{5} & 7 & 3 & 2 & $\mathbf{0.977 \pm 0.017}$ \\
    & \textbf{6} & 4 & 6 & 2 & $\mathbf{0.906 \pm 0.040}$ \\
    & \textbf{7} & 7 & 3 & 2 & $\mathbf{0.977 \pm 0.014}$ \\
    & \textbf{8} & 7 & 3 & 2 & $\mathbf{0.963 \pm 0.025}$ \\
\hline
\end{tabular}
\caption{Per step acceptance, scaling, and rejection statistics for KL based gating
at 0.7 and 0.5 bits per token thresholds. Step 1 is omitted as no scaling 
is applied.}
\label{tab:appendix_kl_gating}
\end{table}

\subsection{Outlier Detection in Per Step KL Divergence}
The per step KL divergence values reported in Tables~1,~3,~and~5 have been processed with outlier detection using a threshold of five standard deviations from the median. Values exceeding this threshold were excluded from the reported statistics. Additionally, steps where the gating mechanism rejected scaling (i.e., could not find a scale factor within the KL divergence budget) were also excluded from per step statistics. Across the reported steps, the sample size after outlier removal and rejection filtering varied from $n=9$ to $n=12$ (out of 12 total runs), indicating that 0--3 data points were excluded per step.

For Tables~2,~4,~and~6, rejected samples were not included in the mean scale calculation, but the remaining scaled and accepted samples were retained for computing aggregate statistics. Consequently, their standard error (SE) estimates also range from $n=9$ to $n=12$, reflecting the variability in sample counts after excluding rejected cases while preserving all valid scaling results.

These extreme outliers are typically caused by individual tokens receiving near zero probability mass under the reference model while being assigned significantly higher probability by the adapted model. This produces disproportionately large contributions to the KL divergence calculation, as KL divergence is computed as:
\begin{equation}
\text{KL}(P \parallel Q) = \sum_{i} P(i) \log \frac{P(i)}{Q(i)},
\end{equation}
where the ratio $P(i)/Q(i)$ becomes extremely large when $Q(i) \approx 0$.

\subsection{Global Drift Measurement}
The total KL divergence used to measure overall model drift also underwent outlier detection, following the same procedure described above. Outlier removal was only necessary for a single instance.

\end{document}